\documentclass[conference]{IEEEtran}
\IEEEoverridecommandlockouts
\usepackage{cite}
\usepackage{amsmath,amssymb,amsfonts}
\usepackage{algorithmic}
\usepackage{graphicx}
\usepackage{textcomp}
\usepackage{xcolor}

\usepackage{multirow}
\usepackage[caption=false,font=footnotesize]{subfig}

\newcommand{\varLatent}{\mathbf{y}}
\newcommand{\varquantLatent}{\hat{\mathbf{y}}}

\newcommand{\varInp}{\mathbf{x}}
\newcommand{\varRec}{\hat{\mathbf{x}}}
\newcommand{\varResi}{\mathbf{r}}

\newcommand{\varFlow}{\mathbf{f}}

\def\BibTeX{{\rm B\kern-.05em{\sc i\kern-.025em b}\kern-.08em
    T\kern-.1667em\lower.7ex\hbox{E}\kern-.125emX}}
\begin{document}

\title{A Combined Deep Learning based End-to-End Video Coding Architecture for YUV Color Space
\thanks{Corresponding authors are A.~K. Singh (ankitesh@qti.qualcomm.com) and H.~E.~Egilmez (hegilmez@qti.qualcomm.com). R.~Pourreza and T.~S. Cohen are with Qualcomm Research, a division of Qualcomm Technologies, Inc.}
}

\author{
\IEEEauthorblockN{Ankitesh K. Singh\textsuperscript{1}, Hilmi E. Egilmez\textsuperscript{1}, Reza Pourreza\textsuperscript{1}, Muhammed Coban\textsuperscript{1}, Marta Karczewicz\textsuperscript{1}, Taco S. Cohen\textsuperscript{2}}
\IEEEauthorblockA{\textsuperscript{1}\textit{Qualcomm Technologies, Inc.}, San Diego, USA \\ \textsuperscript{2}\textit{Qualcomm Technologies Netherlands B.V.}, Amsterdam, Netherlands
}
}

\maketitle

\begin{abstract}
Most of the existing deep learning based end-to-end video coding (DLEC) architectures are designed specifically for RGB color format, yet the video coding standards, including H.264/AVC, H.265/HEVC and H.266/VVC developed over past few decades, have been designed primarily for YUV 4:2:0 format, where the chrominance (U and V) components are subsampled to achieve superior compression performances considering the human visual system. While a broad number of papers on DLEC compare these two distinct coding schemes in RGB domain, it is ideal to have a common evaluation framework in YUV 4:2:0 domain for a more fair comparison. This paper introduces a new DLEC architecture for video coding to effectively support YUV 4:2:0 and compares its performance against the HEVC standard under a common evaluation framework. The experimental results on YUV 4:2:0 video sequences show that the proposed architecture can outperform HEVC in intra-frame coding, however inter-frame coding is not as efficient on contrary to the RGB coding results reported in recent papers.
\end{abstract}


\addtolength{\textfloatsep}{-0.5cm}
\setlength{\abovecaptionskip}{1.5pt}
\setlength{\belowcaptionskip}{-10pt}
\setlength\abovedisplayskip{1.5pt}
\setlength\belowdisplayskip{1.5pt}
\setlength{\belowdisplayskip}{0pt} \setlength{\belowdisplayshortskip}{0pt}
\setlength{\abovedisplayskip}{0pt} \setlength{\abovedisplayshortskip}{0pt}

\section{Introduction}
Deep learning based end-to-end image and video coding (DLEC) architectures are typically designed based on variational autoencoders (VAEs) \cite{Kingma:2014:VAE}, 
where neural networks are trained and tested on sources represented in RGB format. However, most practical video compression systems operate on luminance (luma) and chrominance (chroma) components, represented in a YUV format.
Among various color spaces, YUV 4:2:0 is predominantly adopted as the basic input-output format in many state-of-the-art compression standards, which include the main profiles of High Efficiency Video Coding (HEVC) \cite{Sullivan:12:hevc} and Versatile Video Coding (VVC) \cite{Bross:20:vvc10}\footnote{The main profile of a video coding standard defines the most basic set of instructions that needs be implemented to support the standard in both hardware and software.}. As YUV representation effectively decorrelates the three components, it provides a more compact encoding than coding in RGB domain \cite{Pearlman:2011:DigSigComp}\footnote{Based on our experiments using  HEVC, RGB coding is more than 50\% less efficient than coding the same content in YUV 4:2:0 in terms of bitrate.}. Moreover, for human visual system (HVS), the luma component is much more important than chroma components, since human eye is far less sensitive to color details (chroma components) than to brightness details (luma component) \cite{Pearlman:2011:DigSigComp,Tekalp:2015:DVP}. To take advantage of this HVS behavior, U and V components in YUV 4:2:0 are subsampled without degrading the perceptual quality \cite{Winkler:2001:HVS}. Thus, unlike non-subsampled RGB and YUV 4:4:4 formats, YUV 4:2:0 reduces chroma resolution by four times (in both U and V components) while the luma component is retained at the same resolution. The chroma subsampling in YUV 4:2:0 significantly improves the coding efficiency, since chroma components at a lower resolution are generally coded in fewer bits\footnote{Our experiments with the HEVC standard have shown that coding in YUV 4:2:0 format yields about a 20\% less bitrate on average as compared to coding the same video content in non-subsampled YUV 4:4:4.}.

While coding in YUV 4:2:0 format have greatly benefited traditional codecs, the existing DLEC solutions focus mainly on coding RGB data, and there is very little or no work DLEC designs for YUV 4:2:0 sources in the literature. As demonstrated in our recent work \cite{Egilmez:2021:E2E_intra}, proposing a new DLEC architecture specifically designed to support YUV 4:2:0 for the first time, 
it is ideal for achieving a better coding performance to build a network architecture specialized for YUV 4:2:0 format, whose parameters are trained on a YUV 4:2:0 dataset. Thus, the network has complete and trainable control over luma and chroma fidelity in YUV 4:2:0 color space considering HVS. Besides, such an architecture is essential to fairly compare DLEC solutions against traditional codecs in YUV 4:2:0 domain.

The main goal of this paper to demonstrate the video compression performance of the DLEC scheme by benchmarking against  traditional block-based codecs under a common evaluation framework in YUV 4:2:0 domain. For this purpose, we propose a new DLEC design by combining (i) the DLEC architecture enabling coding in YUV 4:2:0 format from our prior work \cite{Egilmez:2021:E2E_intra} and (ii) the recently developed scale-space-flow (SSF) based inter-frame coding \cite{Agustsson:2020:SSF}. 
Among the DLEC solutions in RGB domain, it has been shown in \cite{Agustsson:2020:SSF} that SSF outperforms existing inter-frame coding DLEC methods \cite{Lu:2019:DVC,Wu:2018:VC_interp} by introducing a \emph{scale} field in addition to traditional \emph{spatial} optical flow for motion compensation. While inter prediction methods in \cite{Agustsson:2020:SSF,Lu:2019:DVC,Wu:2018:VC_interp} are all one-directional (i.e., all inter coded frames are P-frames), Yilmaz and Tekalp \cite{Yilmaz:2020:icip_bidirectoinal} propose a bidirectional inter-frame coding solution with a simple B-frame structure outperforming the P-frame coding in \cite{Lu:2019:DVC}. 
Since HEVC can achieve significant coding gains with more sophisticated,  hierarchical B-frame structures, for a fair comparison, we only focus on P-frame based inter-frame coding and utilize SSF in our DLEC architecture. DLEC approaches for B-frame coding are considered as part of the future work. 

In \cite{Agustsson:2020:SSF}, Agustsson \emph{et al.} further show that SSF achieves competitive compression performance relative to HEVC based on the results obtained using \emph{x265} codec in \emph{ffmpeg} software \cite{software:ffmpeg}. 
In contrast to the line of work focusing on coding RGB data \cite{Agustsson:2020:SSF,Lu:2019:DVC,Wu:2018:VC_interp,Yilmaz:2020:icip_bidirectoinal}, our proposed combined DLEC architecture operates directly on YUV 4:2:0 data, and its performance is compared against the results obtained using the HEVC reference software (HM-16.20). Note that HM is the reference software used as a baseline in standardization activities and many research articles, as it reflects the state-of-the-art compression capabilities of HEVC. However, product-driven compression tools such as \emph{ffmpeg} and \emph{x265} can lead to a significantly lower coding efficiency than HM\footnote{Based on our experiments,  coding with \emph{ffmpeg} under commonly used \emph{medium} presetting for\emph{x265} is about 15\% less efficient than HM in terms of BD-rate \cite{bd_metric}.}. Besides, being closest to our evaluation framework, Sulun and Tekalp \cite{Sulun:2020:Y_only} study neural-network based inter-frame prediction for video compression where both training and testing are performed on Y component only (i.e., gray-level data) by considering the fact that luma component is perceptually more important than chroma components, and the compression performance on luma data is compared against \emph{x265}. 

To the best of our knowledge, this is the first paper in the literature that introduces a complete intra and inter-frame DLEC architecture specialized for YUV 4:2:0 format, where a common evaluation framework is employed to compare it performance over state-of-the-art HEVC defined by HM reference software. 
The present paper can also be viewed as an extension of our prior work on intra-frame coding \cite{Egilmez:2021:E2E_intra} to further support inter-frame coding with SSF.

In the remainder of this paper, Section \ref{sec:proposed_method} presents the proposed combined DLEC architecture. The experimental results are reported and analyzed in Section \ref{sec:results}, and concluding remarks based on our results are drawn in Section \ref{sec:conclusion}.

\begin{figure}[!t]
\centering
    {\includegraphics[width=0.8\columnwidth]{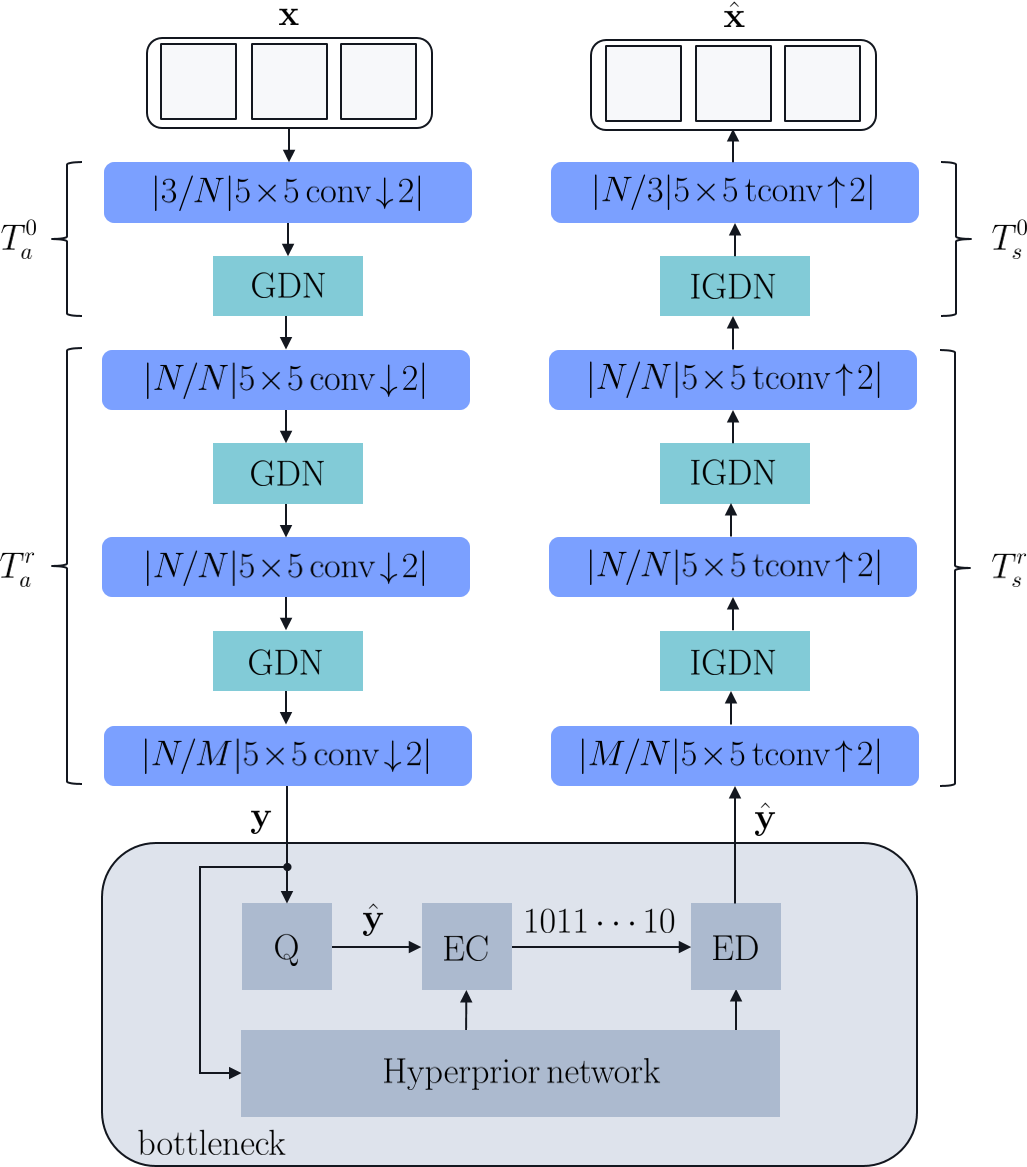}} 
\caption{The transform network architecture used in \cite{MinnenBT:2018:NeurIPS:MSHyper,Agustsson:2020:SSF}, which focus on different bottleneck and hyperprior network designs with the same transform network (i.e., pair of $T_a$ and $T_s$). Moreover, $T_a^0$, $T_a^r$ $T_s^0$ and $T_s^r$ denote the transform sub-networks where 
$\varLatent=T_a(\varInp)=T_a^r(T_a^0(\varInp))$ and $\varRec = T_s(\varquantLatent)=T_s^0(T_s^r(\varquantLatent))$.}    
\label{fig:google_transfrom_network}
\end{figure}

\section{Proposed Combined Architecture}
\label{sec:proposed_method}
\begin{figure}[!t]
\centering
    {\includegraphics[width=0.8\columnwidth]{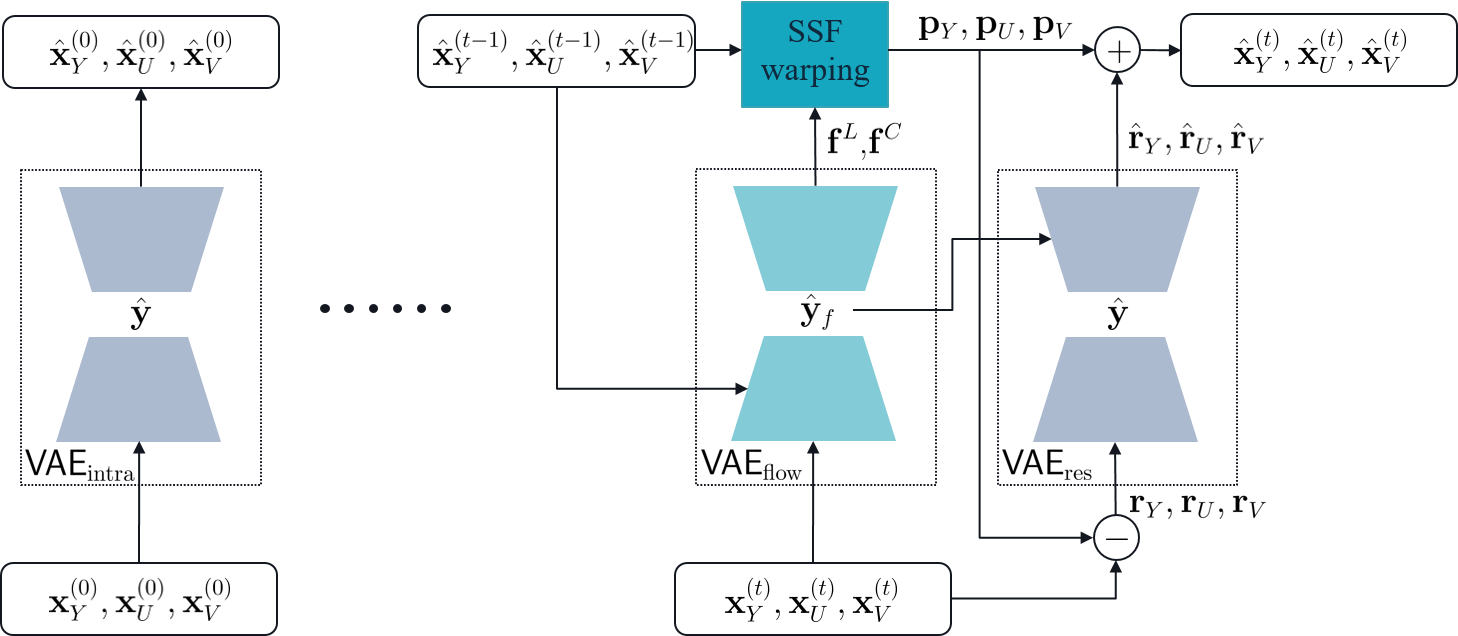}} 
\caption{DLEC architecture for video coding  with SSF \cite{Agustsson:2020:SSF}, where $\mathsf{VAE}_{\text{flow}}$ network is used to learn SSFs for luma ($\varFlow^L$) and chroma ($\varFlow^C$) that are input to the SSF warping module yielding predictions $\mathbf{p}_Y$, $\mathbf{p}_U$ and $\mathbf{p}_V$ subtracted from current YUV input. The resulting residuals are $\mathbf{r}_Y$, $\mathbf{r}_U$ and $\mathbf{r}_V$ coded using $\mathsf{VAE}_{\text{res}}$. Separately, intra-frame coding performed by $\mathsf{VAE}_{\text{intra}}$.}
\label{fig:combined_architecture}
\end{figure}

\begin{figure}[!t]
\centering
{\includegraphics[width=0.9\columnwidth]{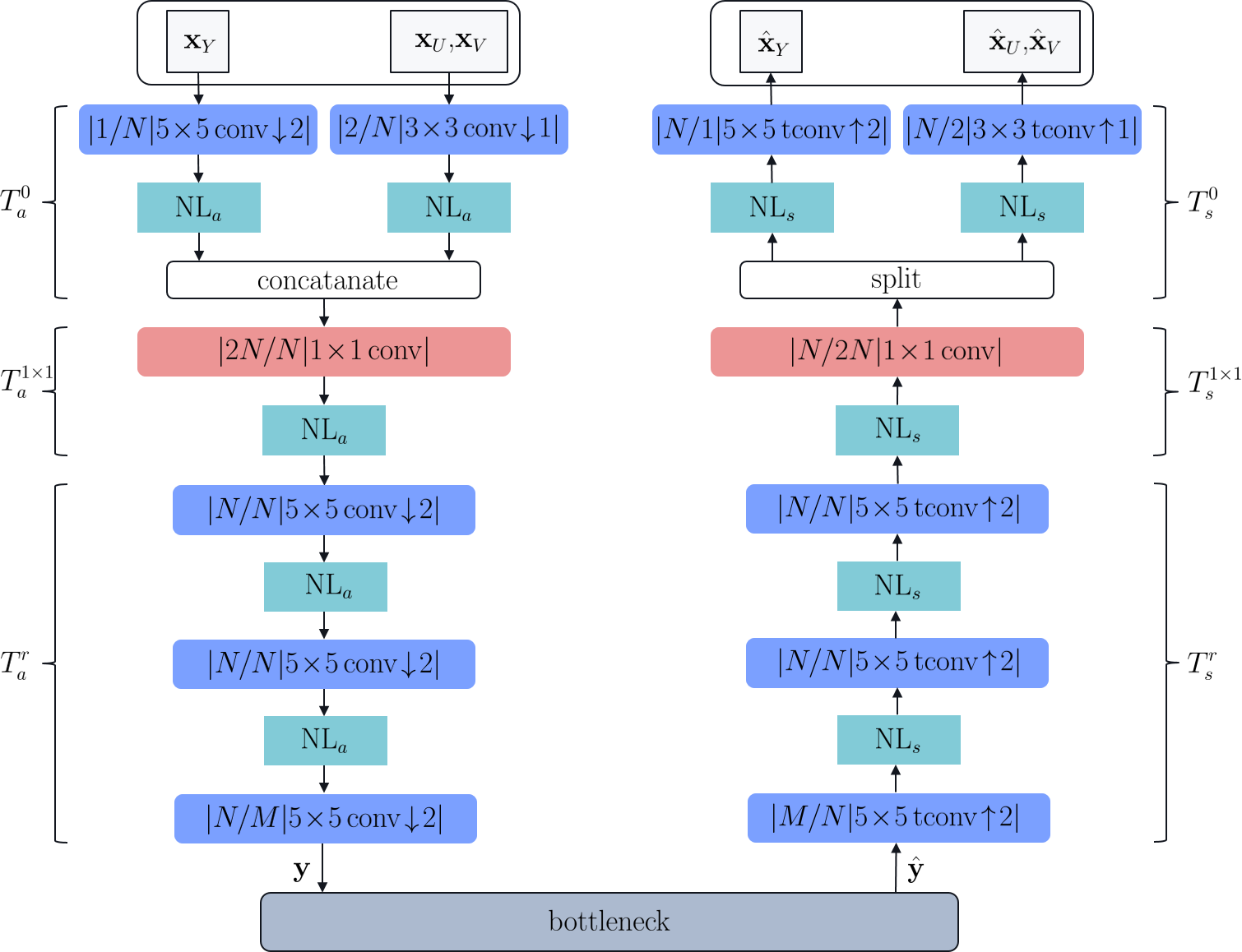}} 
\caption{The architecture proposed in \cite{Egilmez:2021:E2E_intra} to support YUV 4:2:0 input  by integrating (i) the branched network structures $T_a^0$ and $T_s^0$, (ii) additional $1 \times 1$ convolutional layers denoted by $T_a^{1\times1}$ and $T_s^{1\times1}$ and (iii) different choices of nonlinear operators $\mathrm{NL}_a$ and $\mathrm{NL}_s$. The remaining transform network layers denoted by $T_a^r$ and $T_s^r$ follow the same structure in Fig.~\ref{fig:google_transfrom_network}.}    
\label{fig:journal_architecture}
\end{figure}

\begin{figure}[!t]
\centering
{\includegraphics[width=0.7\columnwidth]{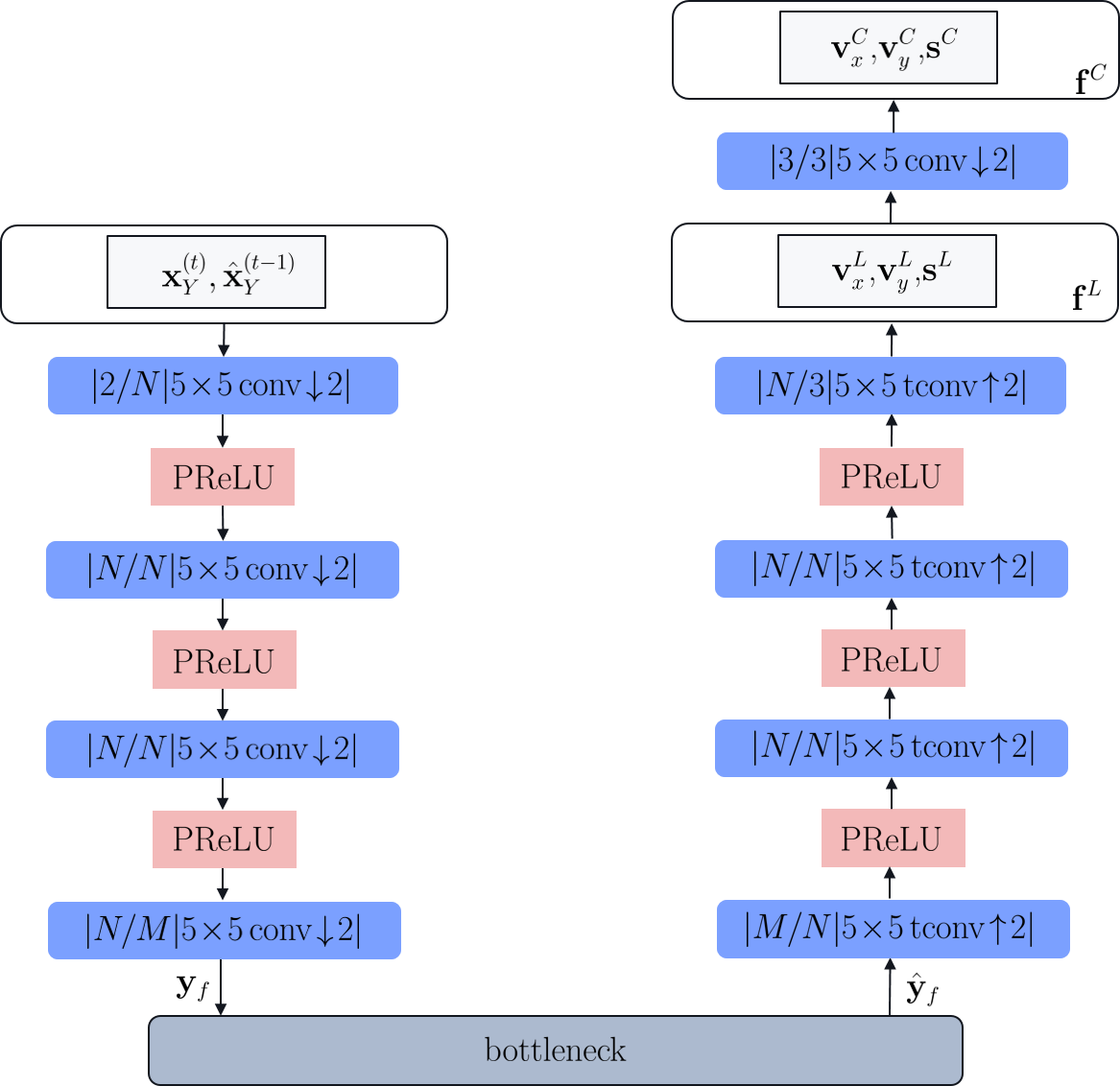}} 
\caption{The architecture used to generate luma and chroma scale-space flows $\varFlow^L$ and $\varFlow^C$ that are input to the SSF warping module in Fig.~\ref{fig:combined_architecture}.}   
\label{fig:flownet}
\end{figure}

Fig.~\ref{fig:combined_architecture} illustrates the overall neural network design for SSF-based video coding, where we propose to incorporate the DLEC architecture in our prior work \cite{Egilmez:2021:E2E_intra} to accommodate YUV 4:2:0 format. 
In \cite{Agustsson:2020:SSF}, variants of the network in Fig.~\ref{fig:google_transfrom_network} are used as the VAEs designed for intra-frame, flow and residual coding, denoted by $\mathsf{VAE}_{\text{intra}}$, $\mathsf{VAE}_{\text{flow}}$ and $\mathsf{VAE}_{\text{res}}$, respectively. 
On the other hand, to effectively support YUV 4:2:0, our combined solution proposes to design $\mathsf{VAE}_{\text{intra}}$ and $\mathsf{VAE}_{\text{res}}$ based on the architecture shown in Fig.~\ref{fig:journal_architecture}, and we further introduce a new $\mathsf{VAE}_{\text{flow}}$ network depicted in Fig.~\ref{fig:flownet} to estimate SSFs for luma and chroma channels (i.e., $\varFlow^L$ and $\varFlow^C$). 

In Fig.~\ref{fig:google_transfrom_network}, the layers denoted by
$|C_{\text{in}}/C_{\text{out}}|K\!\times\!K\,\mathrm{conv}\!\downarrow\!2|$ represent two-dimensional $K\!\times\!K$ convolutions with downsampling by 2 having $C_{\text{in}}$ input and $C_{\text{out}}$ output channels. The corresponding transposed convolutions\footnote{Transposed convolution (i.e., $\mathrm{tconv}$) is also known as deconvolution.} at the decoder is expressed as $|C_{\text{in}}/C_{\text{out}}|K\!\times\!K\,\mathrm{tconv}\!\uparrow\!2|$ with $\uparrow\!2$ standing for upsampling by 2. 
As for nonlinear operators, the analysis transform uses generalized divisive normalization (GDN)~\cite{Balle:2018:GDN} while its synthesis counterpart applies inverted GDN (IGDN) between the convolutions. It has been shown in~\cite{Balle:2018:GDN} that GDNs can considerably improve the RD performance by their cross-channel normalization capability and performs better than some commonly used activation operators including rectified linear unit (ReLU) and leaky ReLU.

The SSF warping essentially applies a trilinear interpolation to generate following inter-frame predictions from learned scale-flow vectors
$\mathbf{f}^L=\left[ \mathbf{v}^L_x,\mathbf{v}^L_y, \mathbf{s}^L \right]$ and  $\mathbf{f}^C=\left[ \mathbf{v}^C_x,\mathbf{v}^C_y, \mathbf{s}^C \right]$ as shown in Fig.~\ref{fig:flownet}. Specifically, luma SSF $\mathbf{f}^L$ is used to generate prediction for Y component ($\mathbf{p}_Y$), and chroma SSF $\mathbf{f}^L$ is for U and V components ($\mathbf{p}_U$ and $\mathbf{p}_V$) as:
\begin{equation}
\begin{split}
\mathbf{p}_Y &:=\mathsf{trilinear}{(\mathbf{x}_Y,\mathbf{f}^L)}  \\
\mathbf{p}_U :=\mathsf{trilinear}&{(\mathbf{x}_U,\mathbf{f}^C)}, \;\;
\mathbf{p}_V :=\mathsf{trilinear}{(\mathbf{x}_V,\mathbf{f}^C)}  
\end{split}
\notag
\end{equation}
such that spatial flow vectors ($\mathbf{v}_x$ and $\mathbf{v}_y$) and scale field ($\mathbf{s}$) for each component (e.g., $\mathbf{v}_x^L$, $\mathbf{v}_y^L$ and $\mathbf{s}^L$ for luma) satisfy the following statements:
\begin{equation}
\begin{split}
\mathbf{p}_Y[x,y] &={\mathbf{x}_Y[x+\mathbf{v}^L_x[x,y],y+\mathbf{v}^L_y[x,y], \mathbf{s}^L[x,y]]}  \\
\mathbf{p}_U[x,y] &={\mathbf{x}_U[x+\mathbf{v}^C_x[x,y],y+\mathbf{v}^C_y[x,y], \mathbf{s}^C[x,y]]} \\
\mathbf{p}_V[x,y] &={\mathbf{x}_V[x+\mathbf{v}^C_x[x,y],y+\mathbf{v}^C_y[x,y], \mathbf{s}^C[x,y]]}  
\end{split}
\notag
\end{equation}
where the scale field $\mathbf{s}$ represents a progressively smoothed versions of the reconstructed frames that is combined together with the spatial displacement/flow information ($\mathbf{v}_x$ and $\mathbf{v}_y$). In \cite{Agustsson:2020:SSF}, $S=5$ scale levels (i.e., blurred version of reconstructed frames) are used. Note that SSF can be viewed as an extension of traditional bilinear warping only using spatial flow information.

In this work, the proposed combined architecture is specifically designed as follows:
\begin{itemize}
    \item $\mathsf{VAE}_{\text{flow}}$ in our design only uses luma component 
    of both the current frame ($\varInp_Y^{(t)}$) and  previously reconstructed frame (($\hat{\varInp}_Y^{(t-1)}$)) to estimate SSFs $\mathbf{f}^L$ and $\mathbf{f}^C$. As shown in Fig.~\ref{fig:flownet}, after learning $\mathbf{f}^L$ a convolutional layer with downsampling is used to learn $\mathbf{f}^C$, as it performs better than directly subsampling $\mathbf{f}^L$ based on our experiments. The number of channels in transform layers and at the bottleneck are selected as $N=192$ and $M=128$, respectively. Note that the proposed flow estimation design is analogous to traditional block-based video coding in the sense that the motion information is typically derived from luma component. 
    \item In our SSF design, we use a Gaussian pyramid with successive filtering and interpolation to generate smoothed versions of reconstructed frames, and $S=3$ scale levels $\mathbf{s}=[0,\sigma_{0}^2,\sigma_{0}^2+ (2\sigma_{0})^2]$ representing Gaussian filter widths, while \cite{Agustsson:2020:SSF} chooses $S=5$. The choice of $S=3$ makes training and inference feasible for high-resolution sequences in terms of memory requirements, and leads to negligible coding performance loss based on our experiments.
    \item $\mathsf{VAE}_{\text{res}}$ is used for coding YUV residual ($\varResi_Y$,$\varResi_U$ and $\varResi_V$) obtained after prediction using SSF warping as shown in Fig.~\ref{fig:combined_architecture}. As for the network architecture, the network shown in Fig.~\ref{fig:journal_architecture} with GDNs from our prior work \cite{Egilmez:2021:E2E_intra}. For this network, both entropy bottleneck and transform layers have 192 channels (i.e., $N=M=192)$.
    \item For intra-frame coding, the architecture shown in Fig.~\ref{fig:journal_architecture} with parameteric ReLU activations is used as the best performing solution from our prior work \cite{Egilmez:2021:E2E_intra}. 
\end{itemize}

\section{Experimental Results}
\label{sec:results}
The proposed combined architecture is implemented, trained and tested on CompressAI framework \cite{begaint:2020:compressai}, which is a deep learning library based on PyTorch for end-to-end image/video compression. 
For entropy coding at the bottleneck, the arithmetic coding engine in \cite{duda:2014:ANS} called asymmetric numeral systems (ANS) is employed. As for the entropy model, the scale hyperprior network in \cite{MinnenBT:2018:NeurIPS:MSHyper} with factorized priors is used as in \cite{Agustsson:2020:SSF} for both residual and flow VAEs, while intra VAE uses the same design in \cite{Egilmez:2021:E2E_intra}. 

For training dataset, we generated YUV 4:2:0 video content by converting the 
448$\times$256 RGB videos in Vimeo-90k Triplet dataset \cite{dataset:vimeo-90k-triplet} into YUV 4:2:0 format using \emph{ffmpeg} software \cite{software:ffmpeg}. 
The resulting YUV 4:2:0 frames (with dimensions 448$\times$256 for luma and 224$\times$128 for chroma components) are used for training, where the batch size is set to 8. Each model is trained for 4 million steps using Adam algorithm \cite{Kingma:2015:iclr:adam} with learning rate set initially $r = 10^{-4}$ for the first two million steps and then set to $r = 10^{-5}$ for the remaining steps.
During training, the following loss function (RD cost) with component-wise weighted mean-squared error (MSE) in the distortion term:
\begin{equation} \label{eqn:loss_initial}
    L = R + \beta \underbrace{(4\,\text{MSE}_Y + \text{MSE}_U + \text{MSE}_V)/6}_{D}  \notag
\end{equation}
where $\beta$ is the parameter used to adjust the trade-off between rate and distortion. $D$ is a weighted combination of MSEs obtained from Y, U and V components (denoted by $\text{MSE}_Y$, $\text{MSE}_U$ and $\text{MSE}_V$, respectively). Note that the weighted terms are proportional to the size of each component (in YUV 4:2:0 format), so that contribution of Y component is 4 times that of U or V. 
For the proposed DLEC approach, four separate models are trained for different $\beta$ parameters ${0.01, 0.025, 0.1, 0.2}$ to evaluate the coding performance at different rate points.

For testing, the class A1 and A2 video sequences in \emph{common test conditions} (CTC) \cite{Bossen:18:ctc} defined for VVC standardization activities are used as our  test dataset, which consists of diverse classes of raw (uncompressed) ultra-high definition (UHD) video sequences in YUV 4:2:0 format. In addition to component-wise PSNR measures, the following combined PSNR metric introduced in \cite{Egilmez:2021:E2E_intra} is reported: 
\begin{equation}
\label{eqn:combined_PSNR}
 \text{Combined PSNR} = (12 \, \text{Y-PSNR} + \text{U-PSNR} + \text{U-PSNR})/{14} \notag
\end{equation}
where the weights for each component (i.e., 12/14, 1/14 and 1/14 for Y, U and V, respectively) are chosen based on our empirical study. 

HEVC benchmark results are obtained by using the HEVC reference software (HM-16.20) under two different configurations:
\begin{itemize}
    \item \emph{low-delay P} (LDP) configuration is used in part of CTC \cite{Bossen:13:hevc_ctc}, where only the first frame is all-intra coded (as I-frame) and all others are coded as P-frames, and
    \item \emph{simple LDP} (SLDP) applies two minor simplifications to LDP configuration for a more fair comparison against the DLEC that (i) allows prediction only from the nearest previously coded frame, (ii) applies quantization parameter (QP) across all frames.
\end{itemize}
Moreover, as the secondary baseline the \emph{x265} encoder in \emph{ffmpeg} is tested under \emph{medium} setting by disabling B-frames.
In both HM and \emph{ffmpeg} tests, class A sequences are coded at five different QPs, that are 42, 37, 32, 27 and 22.

\begin{figure*}[!t]
\centering
    \subfloat[Combined PSNR versus average bitrate\label{fig:combined_classA}]{\includegraphics[width=0.7\columnwidth,trim={0.7cm 0 1.6cm 1.4cm},clip]{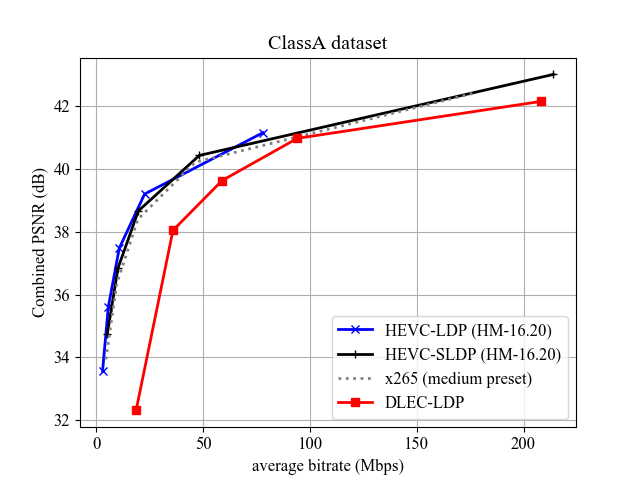}} \qquad
    \subfloat[Y-PSNR versus average bitrate\label{fig:ypsnr_classA}]{\includegraphics[width=0.7\columnwidth,trim={0.7cm 0 1.6cm 1.4cm},clip]{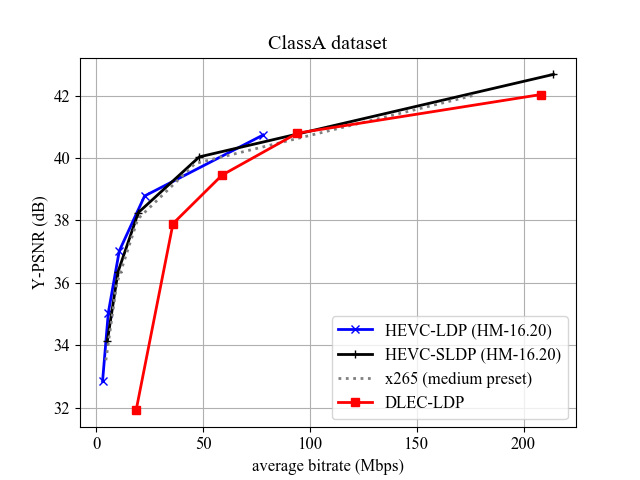}}
    \\ \vspace{-0.2cm}
    \subfloat[U-PSNR versus average bitrate\label{fig:upsnr_classA}]{\includegraphics[width=0.7\columnwidth,trim={0.7cm 0 1.6cm 1.4cm},clip]{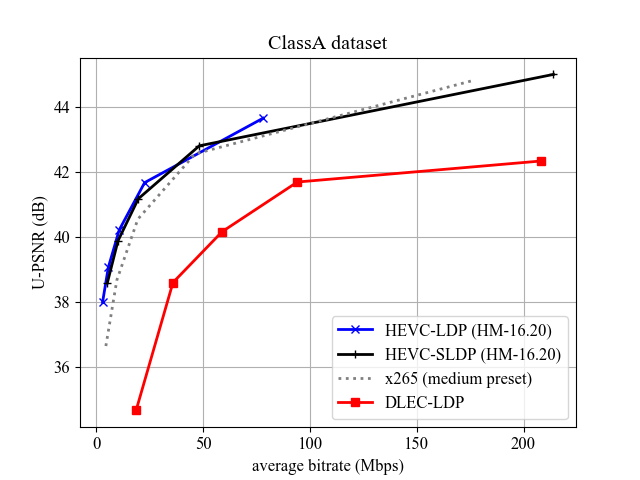}} \qquad
    \subfloat[V-PSNR versus average bitrate\label{fig:vpsnr_classA}]{\includegraphics[width=0.7\columnwidth,trim={0.7cm 0 1.6cm 1.4cm},clip]{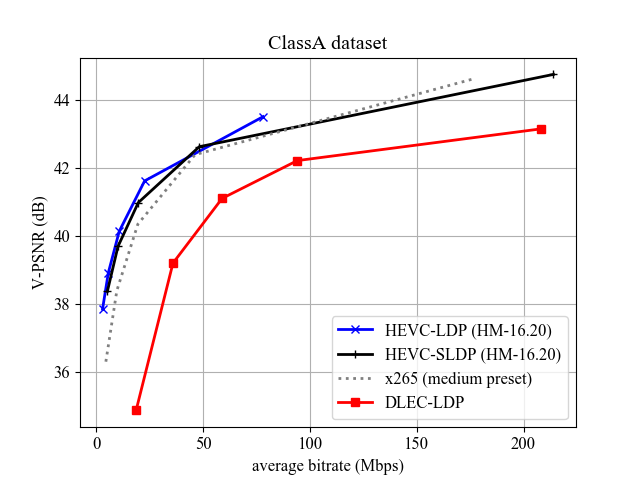}}
\caption{Average compression performance on Class A sequences \cite{Bossen:18:ctc}: HEVC-LDP and HEVC-SLDP results are obtained by using HM under LDP and SLDP configurations, respectively, and x265 (medium preset) stands for results obtained using \emph{x265} codec in \emph{ffmpeg} software under default settings with B-frames disabled. DLEC-LDP denotes the results obtained by the proposed architecture described in Section \ref{sec:proposed_method}.}
\label{fig:inter_classA_results}
\centering
    \subfloat[Combined PSNR versus average bitrate\label{fig:combined_Tango}]{\includegraphics[width=0.7\columnwidth,trim={0.7cm 0 1.6cm 1.4cm},clip]{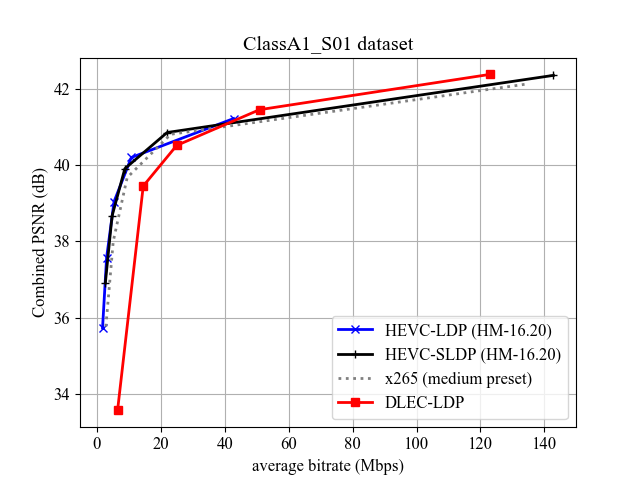}} \qquad
    \subfloat[Y-PSNR versus average bitrate\label{fig:ypsnr_Tango}]{\includegraphics[width=0.7\columnwidth,trim={0.7cm 0 1.6cm 1.4cm},clip]{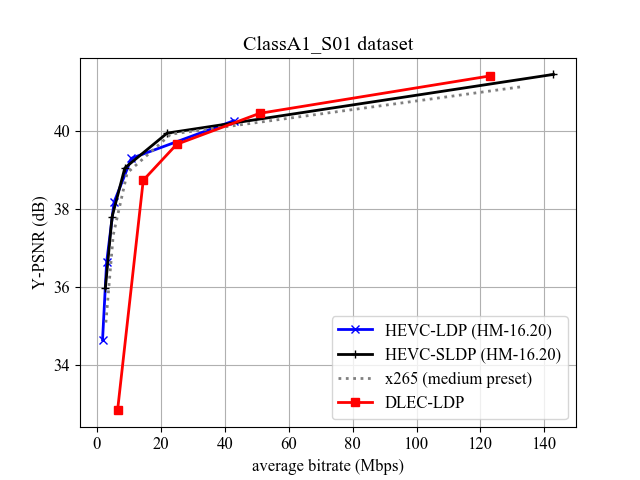}}
    \\  \vspace{-0.2cm}
    \subfloat[U-PSNR versus average bitrate\label{fig:upsnr_Tango}]{\includegraphics[width=0.7\columnwidth,trim={0.7cm 0 1.6cm 1.4cm},clip]{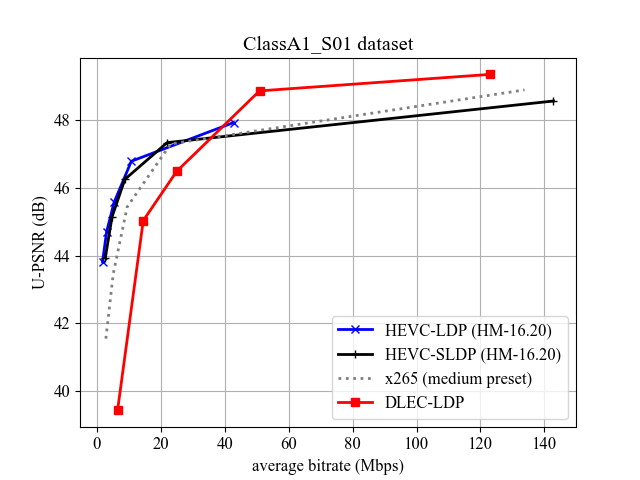}} \qquad
    \subfloat[V-PSNR versus average bitrate\label{fig:vpsnr_Tango}]{\includegraphics[width=0.7\columnwidth,trim={0.7cm 0 1.6cm 1.4cm},clip]{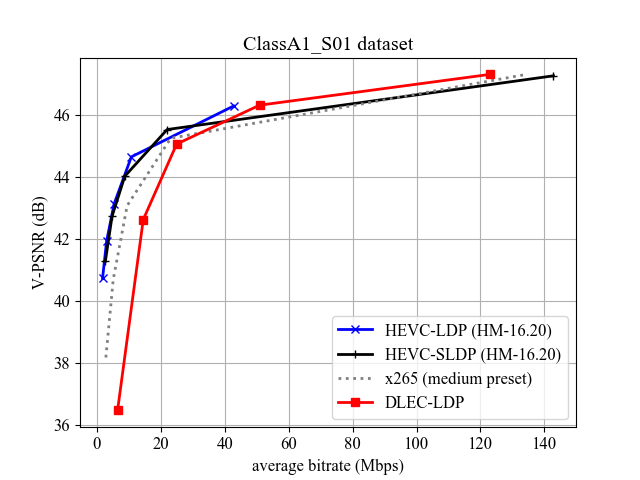}}
\caption{Average compression performance on Tango sequence in \cite{Bossen:18:ctc}: HEVC-LDP and HEVC-SLDP results are obtained by using HM under LDP and SLDP configurations, respectively, and x265 (medium preset) stands for results obtained using \emph{x265} codec in \emph{ffmpeg} software under default settings with B-frames disabled. DLEC-LDP denotes the results obtained by the proposed architecture described in Section \ref{sec:proposed_method}.}
\label{fig:inter_Tango_results}
\end{figure*}

Figures \ref{fig:inter_classA_results} and \ref{fig:inter_Tango_results} compare the YUV 4:2:0 video coding performance of the DLEC by benchmarking against HEVC. Fig.~\ref{fig:inter_classA_results} shows that HEVC significantly outperforms DLEC on average (with more than 50\% BD-rate difference), where the performance gap is much higher in chroma components than in luma. On the other hand, DLEC performs better than HEVC coding of \emph{Tango} sequence (in class A) at high bitrates, while HEVC is superior at low-bitrates as shown in Fig.\ref{fig:inter_Tango_results}.

The performance difference between DLEC and HEVC is purely due to inefficient inter-frame coding of DLEC as our prior work \cite{Egilmez:2021:E2E_intra} shows that the architecture in Fig.~\ref{fig:journal_architecture} achieves about 10\% average BD-rate improvement over HEVC in \emph{all-intra} coding of CTC sequences.
 
\section{Conclusions}
\label{sec:conclusion}
In this paper, we introduced an SSF-based DLEC architecture for video coding to enable coding YUV 4:2:0 data. Our experimental results showed that the state-of-the-art DLEC solutions for video coding are not competitive with the HEVC standard in coding YUV 4:2:0 data, as opposed to the findings in recent studies on coding RGB data. Yet, it is promising that DLEC can outperform HEVC in coding \emph{Tango} sequence, so  the overall coding performance can still be improved with DLEC solutions that better generalize to a wider range of video content. 
Our future work will focus on improving inter-frame performance of DLEC in YUV 4:2:0 domain.

\bibliographystyle{IEEEtran}
\bibliography{refs}

\end{document}